\documentclass[conference]{IEEEtran}
\IEEEoverridecommandlockouts
\usepackage{cite}
\usepackage{amsmath,amssymb,amsfonts}
\usepackage{algorithmic}
\usepackage{graphicx}
\usepackage{textcomp}
\usepackage{xcolor}
\usepackage{url}
\usepackage{multirow}
\usepackage{booktabs}
\usepackage{subfigure}
\usepackage{CJKutf8}
 \usepackage[misc]{ifsym}
\def\BibTeX{{\rm B\kern-.05em{\sc i\kern-.025em b}\kern-.08em
    T\kern-.1667em\lower.7ex\hbox{E}\kern-.125emX}}
\begin{document}

\title{AnchiBERT: A Pre-Trained Model for Ancient Chinese Language Understanding and Generation\\
}

\author{\IEEEauthorblockN{1\textsuperscript{st} Huishuang Tian}
\IEEEauthorblockA{\textit{College of Computer Science} \\
\textit{Sichuan University}\\
Chengdu, China \\
tianhs0075@163.com}
\and
\IEEEauthorblockN{2\textsuperscript{nd} Kexin Yang}
\IEEEauthorblockA{\textit{College of Computer Science} \\
\textit{Sichuan University}\\
Chengdu, China \\
kexinyang0528@gmail.com}
\and
\IEEEauthorblockN{3\textsuperscript{rd} Dayiheng Liu}
\IEEEauthorblockA{\textit{College of Computer Science} \\
\textit{Sichuan University}\\
Chengdu, China \\
losinuris@gmail.com}
\and
\IEEEauthorblockN{4\textsuperscript{th} Jiancheng Lv\textsuperscript{(}\textsuperscript{\Letter}\textsuperscript{)}}
\IEEEauthorblockA{\textit{College of Computer Science} \\
\textit{Sichuan University}\\
Chengdu, China \\
lvjiancheng@scu.edu.cn}

}

\maketitle

\begin{abstract}
Ancient Chinese is the essence of Chinese culture. There are several natural language processing
tasks of ancient Chinese domain, such as ancient-modern Chinese translation, poem generation, and couplet generation. Previous studies usually use the supervised models which deeply rely on parallel data. However, it is difficult to obtain large-scale parallel data of ancient Chinese. In order to make full use of the more easily available monolingual ancient Chinese corpora, we release AnchiBERT, a pre-trained language model based on the architecture of BERT, which is trained on large-scale ancient Chinese corpora. We evaluate AnchiBERT on both language understanding
and generation tasks, including poem classification, ancient-modern Chinese translation, poem generation, and couplet generation. The experimental results show that AnchiBERT outperforms BERT as well as the non-pretrained models and achieves state-of-the-art results in all cases.
\end{abstract}


\section{Introduction}
Ancient Chinese is the written language in ancient China, which has been used for thousands of years. There are large amounts of unlabeled monolingual ancient Chinese text in various forms, such as ancient Chinese articles, poems, and couplets. Investigating ancient Chinese is a meaningful and essential domain. Previous studies have made several attempts on it. For example, \cite{A-MT} trains a Transformer model to translate ancient Chinese into modern Chinese. \cite{couplet} and \cite{couplet-simple} apply an RNN-based model with attention mechanism to generate Chinese couplets. \cite{poem} generates ancient Chinese poems with RNN encoder-decoder framework. These ancient Chinese tasks often employ supervised models, which deeply rely on the scale of parallel datasets. However, those datasets are costly and difficult to obtain due to the requirement for expert annotation. 

In the absence of parallel data, previous studies have proposed pre-trained language models to utilize the large-scale unlabeled corpora to further improve the model performance on NLP tasks, such as ELMo \cite{ELMO}, GPT \cite{GPT}, and BERT \cite{BERT}. These pre-trained models learn universal language representations from large-scale unlabeled corpora with self-supervised objectives, and then are fine-tuned on downstream tasks \cite{lei2021have,lei-etal-2020-examining}. However, these models are trained on general-domain text which has linguistic characteristics shift from ancient Chinese text. The shift between modern Chinese and ancient Chinese is shown in Fig. \ref{ancient-modern sample}. 
 
Therefore, we propose AnchiBERT, a pre-trained language model based on the architecture of BERT, which is trained on the large-scale ancient Chinese corpora. We evaluate the performance of AnchiBERT on both language understanding and generation tasks. Our contributions are as follows:

\begin{figure}
\centering
	\includegraphics[scale=0.5]{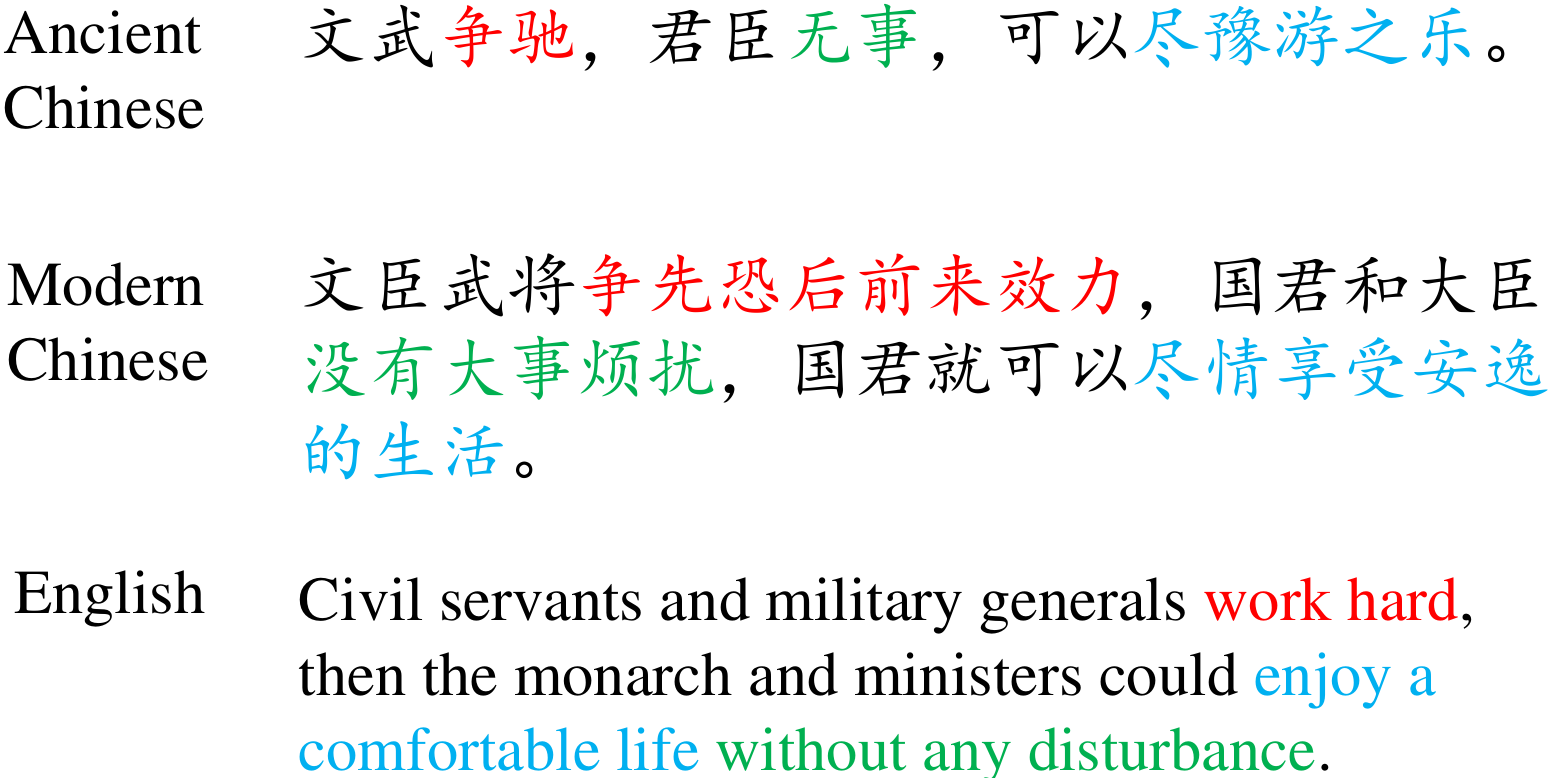}
	\caption{Linguistic characteristics shift between modern Chinese and ancient Chinese.} \label{ancient-modern sample}
\end{figure}

\begin{itemize}
\item To our best knowledge, we propose a first pre-trained language model in ancient Chinese domain, which is trained on the large-scale ancient Chinese corpora we build.

\item We evaluate the performance of AnchiBERT on four ancient Chinese downstream tasks, including both language understanding and language generation tasks. AnchiBERT achieves new state-of-the-art results in all tasks which verify the effectiveness of pre-training strategy in ancient Chinese domain.

\item We propose a complete pipeline to apply pre-trained model into several ancient Chinese domain tasks. We will release our code, pre-trained model, and corpora\footnote{The dataset and model will be available at \url{https://github.com/ttzHome/AnchiBERT}} to facilitate the further research on ancient Chinese domain tasks. 

\end{itemize}

\begin{figure*}
\centering
	\includegraphics[scale=0.57]{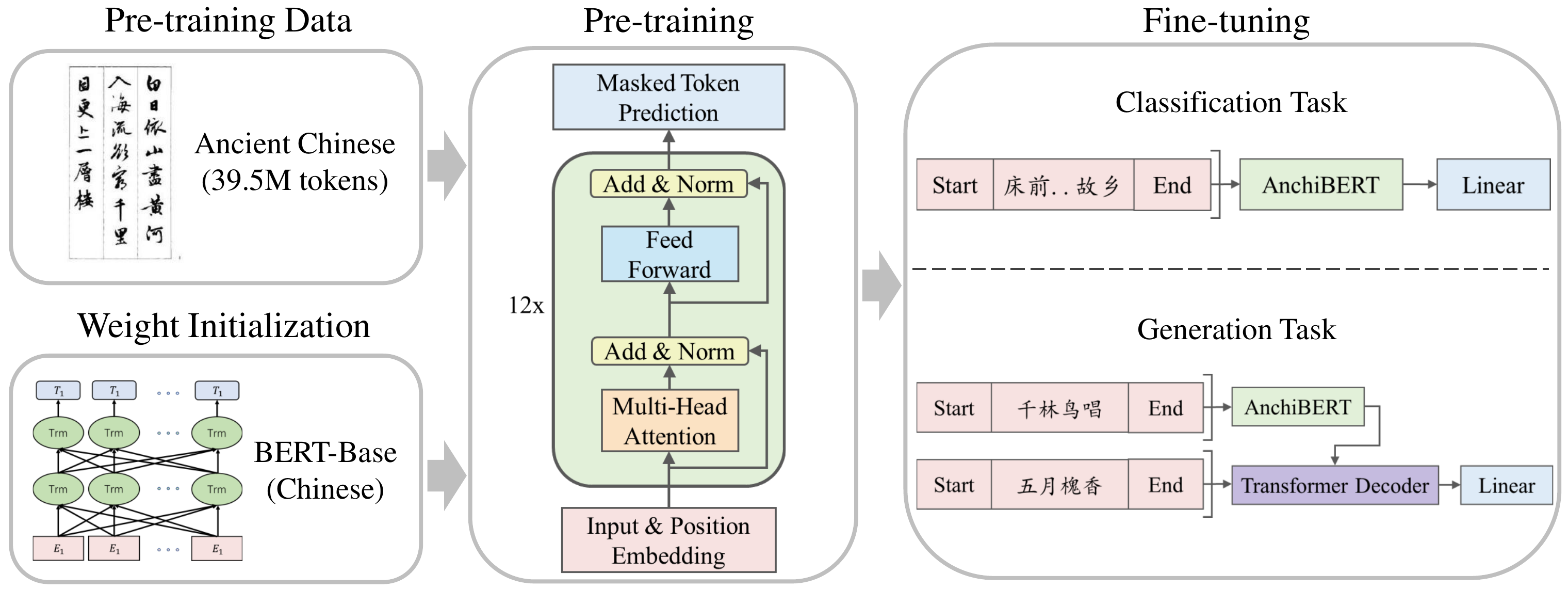}
	\caption{Overview of pre-training and fine-tuning process of AnchiBERT.} \label{pretraining-process}
\end{figure*}

\section{Related Works}

\subsection{Pre-Trained Representations in General}

Pre-training is an effective strategy which is widely used for NLP tasks in recent years. As static representations, Word2Vec \cite{WO2VEC} and GloVe \cite{glove} are the early word-level methods to learn language representations. As dynamic representations, ELMo \cite{ELMO} provides the contextual representations based on a bidirectional language model. ELMo is pre-trained on huge text corpus and can learn better contextualized word embeddings for downstream tasks. GPT \cite{GPT} and BERT \cite{BERT} propose pre-trained Transformer-based model to learn universal language representations by fine-tuning on large-scale corpora. Compared to GPT, BERT is trained on masked token prediction and next sentence prediction tasks. Masked token prediction task extracts bidirectional information instead of unidirectional. Next sentence prediction task predicts if one sentence follows another sentence. Moreover, recent studies propose new pre-trained models, such as XLNet \cite{XLNet}, RoBERTa \cite{RoBERTa}, and ALBERT \cite{ALBERT}, which bring improvements on downstream tasks.

\subsection{Domain-Specific Pre-trained Models}
Several studies propose pre-trained models which adapt to specific domains or tasks. BioBERT \cite{BioBERT} is trained on large-scale biomedical text for biomedical domain tasks. SciBERT \cite{SCIBERT} is trained for scientific domain tasks on biomedical and computer science text, using its own vocabulary (SCIVOCAB). ClinicalBERT \cite{alsentzer-etal-2019-publicly} is proposed due to the need for specialized clinical pre-trained model and is applied to clinical tasks. 
In addition, recent studies also release monolingual pre-trained models for a specific language besides English. FlauBERT \cite{FlauBERT} and CamemBERT \cite{camembert} are trained for French. BERTje \cite{BERTje} and RobBERT \cite{RobBERT} are trained for Dutch. AraBERT \cite{AraBERT} is trained for Arabic language.

\subsection{Ancient Chinese Domain Tasks}
Ancient Chinese domain tasks include translating ancient Chinese into modern Chinese, generating poems, generating couplets, and so on \cite{A-MT,poem-template2,kexin}. For translation, \cite{A-MT} translates ancient Chinese into modern Chinese with a Transformer model. For poem generation, several studies are based on templates and rules\cite{poem-template2,poem-template3,poem-template1}. With the development of deep learning, some approaches generate poems with an encoder-decoder framework\cite{poem-neural3,poem-neural1,poem-neural2,lei-etal-2018-sequicity,lei22}. Moreover, many new model methods are applied to poem generation, such as reinforcement learning \cite{poem-rl} and variational autoencoder \cite{poem-VAE}. For couplet generation, \cite{couplet-smt} uses a statistical machine translation approach. \cite{couplet} and \cite{couplet-simple} apply an RNN-based model with attention mechanism to generate couplets. However, these tasks use limited annotated data and leave the large-scale unlabeled ancient Chinese text behind. We utilize the unlabeled data to train AnchiBERT, a pre-trained model which adapts to ancient Chinese domain. AnchiBERT achieves SOTA results in all downstream tasks.

\section{Method}
\subsection{Model Architecture}
\label{Model Architecture}
AnchiBERT exactly follows the same architecture as BERT \cite{BERT}, using a multi-layer Transformer \cite{Transformer}. AnchiBERT uses the configuration of BERT-base, with 12 layers, the hidden size of 768, and 12 attention heads. The total number of model parameters is about 102M.
\subsection{Pre-Training Data}
\label{Pre-Training Data}
The ancient Chinese corpora used for training AnchiBERT are listed in Table \ref{table1}. The corpora consist of articles, poems and couplets which are written in ancient Chinese, resulting in the corpora size of 39.5M ancient Chinese tokens. Most of our ancient Chinese corpora are written in dynasties of ancient China by many celebrities (about 1000BC-200BC).

We preprocess the raw data crawled from the Internet\footnote{Part of the ancient Chinese text comes from the website \url{http://www.gushiwen.org} and \url{http://wyw.5156edu.com}.}. We first clean the data by removing the useless symbols. Then we convert traditional Chinese characters into simplified characters. Finally, we remove the titles of articles and poems and only leave the bodies.
%
%
%

\renewcommand{\arraystretch}{1.5}
\begin{table}\small
\centering
\caption{\label{table1} Pre-training data used for AnchiBERT. }
\begin{tabular}{cc}
\hline Corpus Type & Number of Tokens \\ \hline
Ancient Chinese Article  & 16.9M  \\
Ancient Chinese Poetry  & 6.7M   \\
Ancient Chinese Couplet  & 15.9M  \\
\hline
\end{tabular}

\end{table}

\subsection{Pre-Training AnchiBERT}
Instead of training from scratch, AnchiBERT continues pre-training based on the BERT-base (Chinese) model\footnote{\url{https://github.com/huggingface/transformers}} on our ancient Chinese corpora, as shown in  Fig. \ref{pretraining-process}. We use masked token prediction task (MLM) to train AnchiBERT. Following \cite{BERT}, given a text sequence $x=\left \{ x_{1},x_{2},...,x_{n} \right \}$ as input, we randomly mask 15\% of the tokens from $x$. During pre-training, 80\% of those selected tokens are replaced with [MASK] token, 10\% are replaced with a random token, and 10\% are unchanged. The training objective is to predict the masked tokens with cross entropy loss. We do not use next sentence prediction (NSP) task because previous work shows this objective does not improve downstream task performance \cite{RoBERTa}.

Following \cite{BERT}, we optimize
the MLM loss using Adam \cite{adam} with a learning rate of 1e-4 and weight decay of 0.01. Due to the limited memory of GPU we train the model with batch size of 15. The maximum sentence length is set to 512 tokens.

We adopt the original tokenization script\footnote{\url{https://github.com/huggingface/transformers/blob/master/src/transformers/tokenization_bert.py}} and tokenize text based on the granularity of Chinese character, where a Chinese character denotes a token. We use the originally released vocabulary in BERT-base (Chinese).

\subsection{Fine-Tuning AnchiBERT}
For ancient Chinese understanding task, we apply a classification layer atop AnchiBERT. For ancient Chinese generation tasks, we use a Transformer-based encoder-decoder framework, which employs AnchiBERT as encoder and uses a transformer decoder with random initialization parameters. Details can be found in \S~\ref{Fine-tuning}.

\section{Experiments}
\label{Experiment}
In this section, we first describe the pre-training details of AnchiBERT, and then introduce the task objective, dataset, settings, baselines, and metrics of each downstream task.
\subsection{AnchiBERT Pre-training}
\label{sect:Task}

AnchiBERT continues pre-training from BERT-base (Chinese) on our ancient Chinese corpora rather than from scratch. AnchiBERT follows the same configuration as BERT-base. Details of model configuration and pre-training data are in \S~\ref{Model Architecture} and \S~\ref{Pre-Training Data} respectively.

During pre-training, we set the maximum sentence length of 512 tokens to train the model on masked token prediction task with Adam optimizer. The batch size is 15 and
training steps are 250K. We use 3 RTX 2080ti GPUs for training. AnchiBERT training takes about 3 days. Our code is implemented based on the Pytorch-Transformers library released by huggingface\footnote{\url{https://github.com/huggingface/transformers}} \cite{Wolf2019HuggingFacesTS}. 

\renewcommand{\arraystretch}{1.5}
\begin{table}
\centering
\caption{\label{table2}  Train/dev/test dataset sizes of each task.}
  \fontsize{10}{9}\selectfont 
\begin{tabular}{cc}
\hline Task & Data(train$/$dev$/$test) \\ \hline
PTC &2.8K$/$0.2K$/$0.2K  \\
AMCT  &1.0M$/$125.7K$/$100.6K\\
CPG  &0.22M$/$5.4K$/$5.4K  \\
CCG  &0.77M$/$4.0K$/$4.0K  \\
\hline
\end{tabular}

\end{table}

\subsection{AnchiBERT Fine-tuning}
\label{Fine-tuning}
\subsubsection{Poem Topic Classification (PTC)}
\label{Fine-tuning on Poem Topic Classification}
Given a poem, the objective of Poem Topic Classification (PTC) task is to obtain the corresponding literary topic. We fine-tune and evaluate AnchiBERT on a publicly released dataset\footnote{\url{https://github.com/shuizhonghaitong/classification\_GAT/tree/master/data}}. The dataset contains 3.2K four-line classical Chinese poems combined with titles and keywords, and each poem has one annotated literary topic (e.g., farewell poem, warfare poem). Details of data splits are shown in Table \ref{table2}.

For training settings, we feed the final hidden vector corresponding to [CLS] token into a classification layer to obtain the topic label, as Fig. \ref{pretraining-process} shows. The input is the body of a poem and output is the corresponding topic label. We apply a batch size of 24 and use Adam optimizer with a learning rate of 5e-5. The dropout rate is always 0.1. The number of training epoch is around 5.

We compare our AnchiBERT with the following baselines:
\begin{enumerate}
    \item Std-Transformer: Std-Transformer is a standard Transformer encoder following the same architecture and configuration as official BERT-base (Chinese), such as the number of layers and hidden size. The vocabulary is the same as well. However, the training weights are randomly initialized instead of pre-trained.
    \item BERT-Base: We choose the pre-trained weights of official version BERT-base (Chinese) \cite{BERT} to initialize BERT-Base. We adopt the original vocabulary.
\end{enumerate}
For automatic evaluation metric, we evaluate models on classification accuracy.

\subsubsection{Ancient-Modern Chinese Translation (AMCT)}
\label{Fine-tuning on Ancient-Modern Chinese Translation}
Ancient-Modern Chinese Translation (AMCT) task translates ancient Chinese sentences into modern Chinese, because ancient Chinese is difficult for modern people to understand. We conduct experiments on ancient-modern Chinese dataset \cite{A-MT}. This dataset contains 1.2M aligned ancient-modern Chinese sentence pairs, with ancient Chinese sentence as input and modern Chinese as target.

For training settings, this task is based on encoder-decoder framework. As Fig. \ref{pretraining-process} shows, we initialize the encoder with AnchiBERT and use a Transformer-based decoder, which is randomly initialized. Following the framework of Transformer, our decoder generates text conditioned on encoder hidden representations through multi-head attention. The training objective is to minimize
the negative log likelihood of the generated text.

The training batch size and the layer number of decoder is 30 and 4, respectively. We use the same optimizer as Transformer, with $\beta_{1}$ = 0.9, $\beta_{2}$ = 0.98, $\epsilon$ = 1e-9 and a linear warmup over 4000 steps. The dropout rate is 0.1. We choose the best number of epoch on the Dev set.

We compare our AnchiBERT with the following baselines:
\begin{enumerate}
    \item Transformer-A: Transformer-A \cite{A-MT} is a Transformer model with augmented data of ancient-modern Chinese pairs.
    \item Std-Transformer: Std-Transformer follows the framework of Transformer, with an encoder identical to Std-Transformer in  \S~\ref{Fine-tuning on Poem Topic Classification} and a randomly initialized decoder.
    \item BERT-Base: BERT-Base follows the framework of Transformer, with an encoder identical to BERT-Base in  \S~\ref{Fine-tuning on Poem Topic Classification} and a randomly initialized decoder.
\end{enumerate}
For automatic evaluation metric, we adapt BLEU evaluation \cite{BLEU} which compares the quality of generated sentences with the ground truth. We apply BLEU-4 in this task. 

We also include human evaluation for generation tasks because the above automatic evaluation metric has some flaws. For example, given an ancient Chinese sentence, there is only one ground truth. But in fact there are more than one appropriate ways of expression for modern Chinese. Thus we follow the evaluation standards in \cite{couplet}, and invite 10 evaluators to rank the generations in two aspects: syntactic and semantic. As for syntactic, evaluators evaluate whether the composition of translated modern Chinese is complete. As for semantic, evaluators consider whether the generated sentences are coherent and fluent. The score is assigned with 0 and 1, with 1 meaning good.

\subsubsection{Chinese Poem Generation (CPG)}
In Chinese Poem Generation (CPG) task, we implement two experimental settings. The first task is to generate the last two lines of a poem from the first two lines (2-2), the second task is to generate the last three lines from the first line (1-3). These four lines of a poem should match each other by following the syntactic and semantic rules in ancient Chinese poems. 
We use another publicly available poetry dataset\footnote{\url{https://github.com/chinese-poetry/chinese-poetry}} for experiment, which contains 0.23M four-line classical Chinese poems.

For training settings, this task uses the same encoder-decoder framework and loss function as AMCT described in \S~\ref{Fine-tuning on Ancient-Modern Chinese Translation}. We apply a batch size of 80 and a 2-layer randomly initialized decoder. We use the same optimizer as AMCT in \S~\ref{Fine-tuning on Ancient-Modern Chinese Translation}. We choose the best number of epoch on the Dev set.

We compare our AnchiBERT with the following baselines:
\begin{enumerate}
    \item Std-Transformer: Std-Transformer follows the framework of Transformer, with an encoder identical to Std-Transformer in  \S~\ref{Fine-tuning on Poem Topic Classification} and a randomly initialized decoder.
    \item BERT-Base: BERT-Base follows the framework of Transformer, with an encoder identical to BERT-Base in  \S~\ref{Fine-tuning on Poem Topic Classification} and a randomly initialized decoder.
\end{enumerate}
For automatic evaluation metric, we use BLEU-4 in this task. Meanwhile, we follow the human metric in \S~\ref{Fine-tuning on Ancient-Modern Chinese Translation} to evaluate the generated poems in syntactic and semantic. Especially, for syntactic, evaluators consider whether the generated poem sentences conform to the length and rhyming rules.

\renewcommand{\arraystretch}{1.5}
\begin{table*}
	\centering
	\caption{Human evaluation results of generation tasks}\label{human eval}
	\fontsize{9}{9}\selectfont 
	\begin{tabular}{c|cc|cc|cc|cc|c}
	\hline
	\multirow{2}{*}{Model}&
	\multicolumn{2}{c|}{AMCT}&\multicolumn{2}{c|}{CPG (2-2)}&\multicolumn{2}{c|}{CPG (1-3)}&\multicolumn{2}{c|}{CCG}&\multirow{2}{*}{Average} \\
	&Syntactic&Semantic&Syntactic&Semantic&Syntactic&Semantic&Syntactic&Semantic \\
	\hline
	Std-Transformer&0.63&0.58&0.69&0.60&0.63&0.52&0.61&0.59&0.61 \\
	BERT-Base&0.69&0.61&0.72&0.64&0.67&0.54&0.63&0.62&0.64 \\
	AnchiBERT&{\bf 0.71}&{\bf 0.62}&{\bf 0.73}&{\bf 0.65}&{\bf 0.69}&{\bf 0.55}&{\bf 0.65}&{\bf 0.63}&{\bf 0.65} \\
	\hline
	\end{tabular}
	
\end{table*}

\renewcommand{\arraystretch}{1.5} 
\begin{table}[h]  
  
  \centering  
\caption{\label{table-result1}
Evaluation results on AMCT and CPG tasks 
}
  \fontsize{10}{9}\selectfont  
  \label{tab:performance_comparison}  
    \begin{tabular}{ccc}
\hline
{Task}&
{Model}&  
{BLEU-4}\\  
\hline 
    \multirow{4}{*}{AMCT}&
    Transformer-A&27.16\cr  
    &Std-Transformer&27.80\cr 
    &BERT-Base&28.89\cr
    &AnchiBERT&{\bf 31.22}\cr  
\hline
    \multirow{3}{*}{CPG (2-2)}&
    Std-Transformer&27.47\cr 
    &BERT-Base&29.82\cr
    &AnchiBERT&{\bf 30.08}\cr 
    \cmidrule(lr){2-3}
    \multirow{3}{*}{CPG (1-3)}&
    Std-Transformer\footnotemark[9]&19.52\cr 
    &BERT-Base&21.63\cr
    &AnchiBERT&{\bf 22.10}\cr  
\hline
    \end{tabular}

\end{table}

\renewcommand{\arraystretch}{1.5} 
\begin{table}[h]  
  
  \centering  
    \caption{\label{table-result-ccg}
Evaluation results on CCG task
}
  \fontsize{10}{9}\selectfont  
  \label{tab:performance_comparison}  
    \begin{tabular}{ccc}
\hline
{Task}&
{Model}&  
{BLEU-2}\\  
\hline
    \multirow{6}{*}{CCG}&
    LSTM&10.18\cr  
    &Seq2Seq&19.46\cr 
    &SeqGAN&10.23\cr
    &NCM&20.55\cr  
    &Std-Transformer&27.14\cr 
    &BERT-Base&33.01\cr
    &AnchiBERT&{\bf 33.37}\cr  
\hline
    \end{tabular}

\end{table}

\subsubsection{Chinese Couplet Generation (CCG)}
Chinese Couplet Generation (CCG) task generates the second sentence (namely a subsequent clause) of couplet, given the first sentence (namely an antecedent clause) of couplet. We conduct this experiment on a publicly available couplet dataset\footnote{\url{https://github.com/wb14123/couplet-dataset}}, which contains 0.77M couplet pairs.

For training settings, we use the same model architecture and loss function described in \S~\ref{Fine-tuning on Ancient-Modern Chinese Translation}. The batch size is 80 and the layer number of decoder is 4. We use the same optimizer in \S~\ref{Fine-tuning on Ancient-Modern Chinese Translation} and fine-tune for around 60 epochs. 

We compare our AnchiBERT with the following baselines:
\begin{enumerate}
    \item RNN-based Models: We first implement the basic LSTM and Seq2Seq model, which has been successfully used in a lot of generation tasks like dialogue systems \cite{lei2018sequicity, jin2018explicit, liao2020rethinking}. We also include SeqGAN model \cite{SeqGAN}, which applies reinforcement learning into Generative Adversarial Net (GAN) to solve the problems in generating discrete sequence tokens. Furthermore, NCM \cite{couplet} is an RNN-based Seq2Seq model incorporating the attention mechanism. NCM also includes a polishing schema, which generates a draft first and then refines the wordings.
    \item Std-Transformer: Std-Transformer follows the framework of Transformer, with an encoder identical to Std-Transformer in  \S~\ref{Fine-tuning on Poem Topic Classification} and a randomly initialized decoder.
    \item BERT-Base: BERT-Base follows the framework of Transformer, with an encoder identical to BERT-Base in  \S~\ref{Fine-tuning on Poem Topic Classification} and a randomly initialized decoder.
\end{enumerate}
For automatic evaluation metric, because the generated couplet sentences are often less than 10 tokens, we use BLEU-2 in CCG task. Meanwhile, we use the human evaluation metric in \S~\ref{Fine-tuning on Ancient-Modern Chinese Translation} to evaluate couplet in syntactic and semantic. For syntactic, the generated subsequent clauses should conform to the length and pattern rules.

\begin{figure*}[h]
 \centering 
 \subfigure[]{
  \label{Fig.sub.1}
  \includegraphics[width=0.48\textwidth]{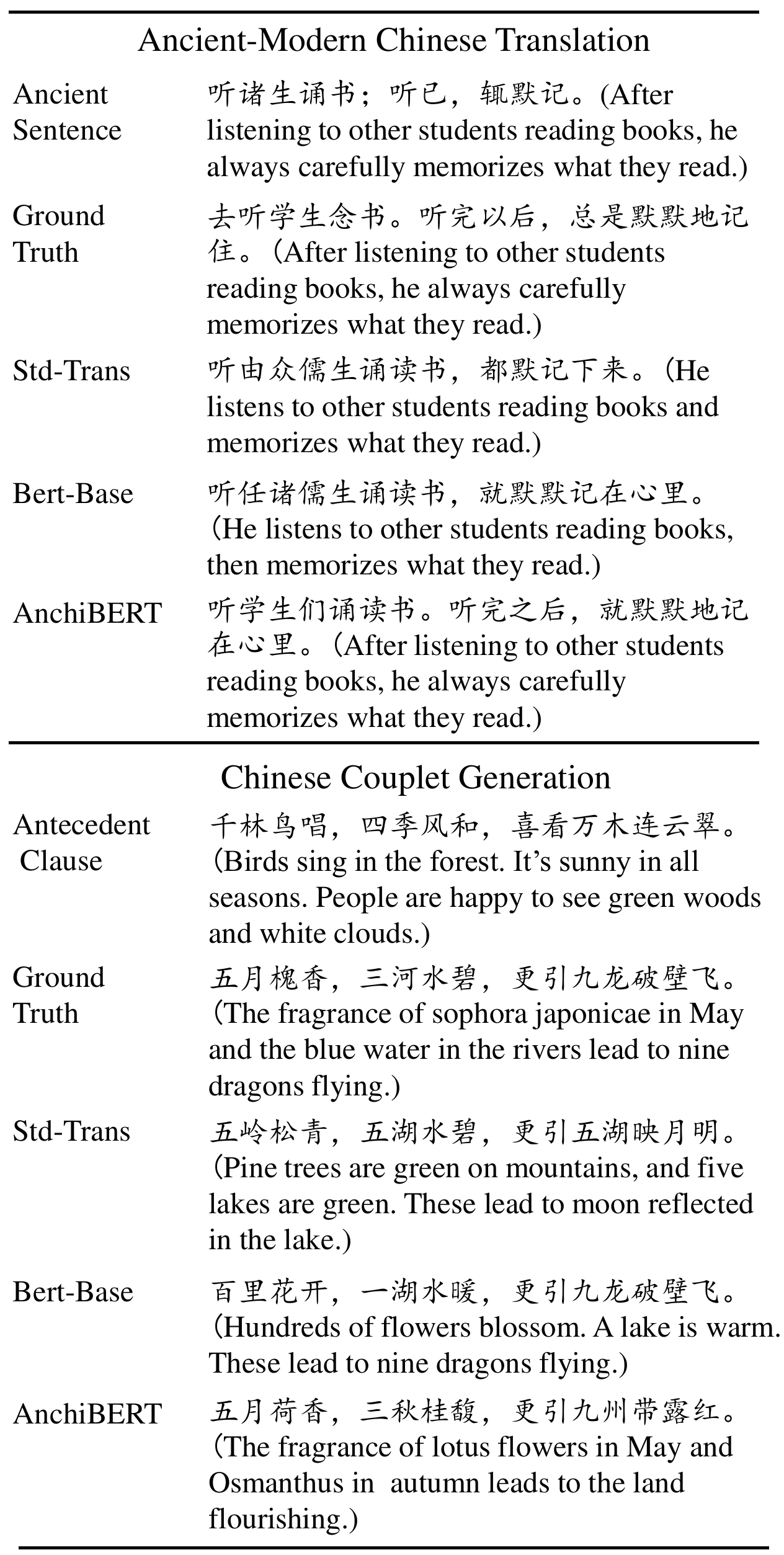}}
 \hspace{.01in} 
 \subfigure[]{
  \label{Fig.sub.2}
  \includegraphics[width=0.48\textwidth]{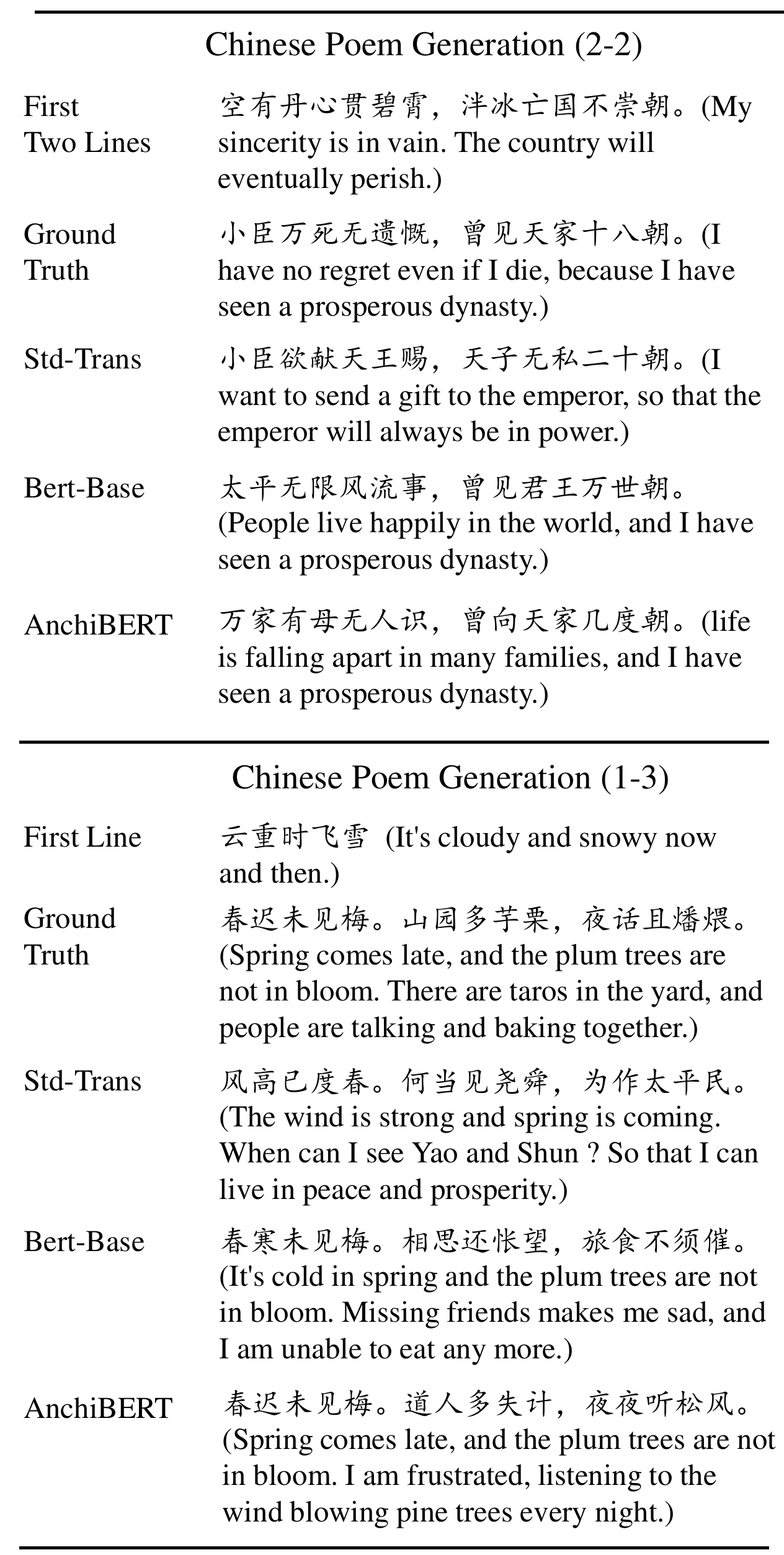}}
\centering
\caption{Sample comparison of generation tasks. 'Std-Trans' in the figure is short for Std-Transformer.}
\label{samples}
\end{figure*}

\section{Results}

\label{sec:Result}
The experiment results are shown in tables above. Generally, we find that AnchiBERT outperforms BERT-Base as well as the non-pretrained models on all ancient Chinese domain tasks. AnchiBERT also achieves new SOTA results in all cases.

\subsection{Automatic Evaluation Results}
The accuracy (the higher the better) is shown in Table \ref{table-result2} and BLEU (the higher the better) results are shown in Table \ref{table-result1}  and Table \ref{table-result-ccg} respectively.

\label{task results }
\paragraph{Poem Topic Classification (PTC)}
Table \ref{table-result2} shows AnchiBERT achieves the SOTA result in Poem Topic Classification task. AnchiBERT improves accuracy by 6.99 over BERT-Base and 12.34 over Std-Transformer. Because the scale of this task dataset is very small, the result illustrates pre-training, especially domain-specific pre-training can significantly improve performance on low-resource task. 

\footnotetext[9]{The performance of Std-Transformer (12 layers of encoder) is extremely poor for CPG (1-3), so we train a randomly initialized Transformer (6 layers of encoder) for this experimental setting and present the best result.}

\paragraph{Ancient-Modern Chinese Translation (AMCT)}
Table \ref{table-result1} shows AnchiBERT outperforms all the baseline models in Ancient-Modern Chinese Translation task. AnchiBERT raises the BLEU score by 2.33 points over BERT-Base and 3.42 over Std-Transformer, which demonstrates the effectiveness of domain-specific pre-training in language generation task.

\paragraph{Chinese Poem Generation (CPG)}
As Table \ref{table-result1} shows, we implement two experimental settings for CPG task, including generating the last two sentences from the first two sentences (2-2) and generating the last three sentences from the first sentence (1-3). AnchiBERT improves performance over two variants (BERT-Base and Std-Transformer) in both experimental settings. In CPG (2-2), AnchiBERT reaches a slightly higher score by 0.26 than BERT-Base and +2.62 than Std-Transformer. In CPG (1-3), AnchiBERT reaches +0.47 over BERT-Base and +2.58 over Std-Transformer. 

\begin{table}
\centering
\caption{Results on Poem Topic Classification task}\label{table-result2}
\fontsize{10}{9}\selectfont 
\begin{tabular}{cc}
\hline
Model & Accuracy (\%)\\
\hline
Std-Transformer& 69.96 \\
BERT-Base & 75.31 \\
AnchiBERT & \textbf{82.30} \\
\hline
\end{tabular}

\end{table}

\paragraph{Chinese Couplet Generation (CCG)}
Table \ref{table-result-ccg} shows evaluation result of CCG task, and in this task we apply BLEU-2 as evaluation metric. AnchiBERT outperforms all of the non-pretrained baseline models and two variants (+0.36 over BERT-Base and +6.23 over Std-Transformer). Note that the task-specific model NCM performs better than general model Std-Transformer, which demonstrates the need for task-specific model architectures. However, the pre-trained models (AnchiBERT and BERT-Base) outperform NCM. This illustrates that sometimes simple pre-trained model is better than complex model architectures.
\\
\\
Our goal of proposing AnchiBERT is to confirm the performance of pre-training strategy in ancient Chinese domain. As we expect, all pre-trained models (AnchiBERT and BERT-Base) perform better than non-pretrained baselines. Meanwhile, AnchiBERT achieves new SOTA results on all ancient Chinese domain tasks.

\subsection{Human Evaluation Results}
\label{sect:discussion}
Table \ref{human eval} reports the human evaluation results on generation tasks. We only compare with BERT variants (Std-Transformer and BERT-Base) because we focus on the effectiveness of domain-specific pre-training. For each experiment, we collect 20 generations respectively. We invite 10 evaluators who are proficient in
Chinese literature. 

In general, the average results demonstrate our model AnchiBERT outperforms all variants. The syntactic scores of our pre-trained AnchiBERT show that although no templates or rules (such as rhythm and length for poem) are set in the AnchiBERT model explicitly, the model can automatically generate text conforming to these grammatical rules. The semantic scores indicate that AnchiBERT learns semantic rules during pre-training, so in downstream tasks AnchiBERT can generate more coherent text across sentences. Note that BERT-Base achieves similar scores with AnchiBERT, which demonstrates pre-training on general-domain text is efficient as well.

\subsection{Samples Analysis}
\label{sect:Samples and Analysis}
Fig. \ref{samples} shows some samples of ancient Chinese translation, poem generation and couplet generation. In the generation tasks, we observe that the inability of Std-Transformer to learn language representation leads to the lack of coherence in generated sentences. BERT-Base learns representation from modern Chinese corpus, so it performs slightly worse for ancient Chinese. AnchiBERT is able to generate ancient Chinese sentences which is coherent and meaningful. 

For example, in Ancient-Modern Chinese Translation task, ancient sentence `\begin{CJK}{UTF8}{gkai}听已\end{CJK}' (after listening) is translated into `\begin{CJK}{UTF8}{gkai}听完以后\end{CJK}' (after listening). However, Std-Transformer and BERT-Base ignore this sentence, whereas AnchiBERT makes the translation. In Chinese Poem Generation (2-2), the original ground truth describes the patriotism of the author. However, the generated sentences of Std-Transformer do not have this meaning. Meanwhile, the first generated sentence of BERT-Base describes the life of ordinary people, which has a semantic shift from the ground truth. AnchiBERT generates sentences which express the heavy atmosphere and the expectations for a prosperous dynasty and fit the poem topic well.

%
%
%

\subsection{Discussion}
\label{sect:discussion}
We observe that pre-training is an effective strategy in ancient Chinese domain, not only in language understanding but also in language generation tasks. On automatic evaluation, AnchiBERT performs better than BERT-base in all ancient Chinese domain tasks, and significantly outperforms the non-pretrained models. Human evaluators also think that AnchiBERT is able to generate text which follows grammatical rules better and is more fluent for people to read.

\section{Conclusion}
\label{sec:Result}
In this paper, we release AnchiBERT, the first pre-trained language model in ancient Chinese domain to the best of our knowledge. AnchiBERT is based on BERT and trained on ancient Chinese corpora. We evaluate AnchiBERT on downstream language understanding and generation tasks, which achieves state-of-the-art performance.

There are some directions for future research. First, find a suitable learning objective during pre-training in ancient Chinese domain. Then, find more ancient Chinese data and construct an ancient Chinese domain vocabulary to train AnchiBERT.

\bibliographystyle{IEEEtran}

\begin{thebibliography}{10}
\providecommand{\url}[1]{#1}
\csname url@samestyle\endcsname
\providecommand{\newblock}{\relax}
\providecommand{\bibinfo}[2]{#2}
\providecommand{\BIBentrySTDinterwordspacing}{\spaceskip=0pt\relax}
\providecommand{\BIBentryALTinterwordstretchfactor}{4}
\providecommand{\BIBentryALTinterwordspacing}{\spaceskip=\fontdimen2\font plus
\BIBentryALTinterwordstretchfactor\fontdimen3\font minus
  \fontdimen4\font\relax}
\providecommand{\BIBforeignlanguage}[2]{{%
\expandafter\ifx\csname l@#1\endcsname\relax
\typeout{** WARNING: IEEEtran.bst: No hyphenation pattern has been}%
\typeout{** loaded for the language `#1'. Using the pattern for}%
\typeout{** the default language instead.}%
\else
\language=\csname l@#1\endcsname
\fi
#2}}
\providecommand{\BIBdecl}{\relax}
\BIBdecl

\bibitem{A-MT}
\BIBentryALTinterwordspacing
D.~Liu, K.~Yang, Q.~Qu, and J.~Lv, ``Ancient-modern chinese translation with a
  new large training dataset,'' \emph{{ACM} Trans. Asian Low Resour. Lang. Inf.
  Process.}, vol.~19, no.~1, pp. 6:1--6:13, 2020. [Online]. Available:
  \url{https://doi.org/10.1145/3325887}
\BIBentrySTDinterwordspacing

\bibitem{couplet}
\BIBentryALTinterwordspacing
R.~Yan, C.~Li, X.~Hu, and M.~Zhang, ``Chinese couplet generation with neural
  network structures,'' in \emph{Proceedings of the 54th Annual Meeting of the
  Association for Computational Linguistics, {ACL} 2016, August 7-12, 2016,
  Berlin, Germany, Volume 1: Long Papers}.\hskip 1em plus 0.5em minus
  0.4em\relax The Association for Computer Linguistics, 2016. [Online].
  Available: \url{https://doi.org/10.18653/v1/p16-1222}
\BIBentrySTDinterwordspacing

\bibitem{couplet-simple}
\BIBentryALTinterwordspacing
S.~Yuan, L.~Zhong, L.~Li, and R.~Zhang, ``Automatic generation of chinese
  couplets with attention based encoder-decoder model,'' in \emph{2nd {IEEE}
  Conference on Multimedia Information Processing and Retrieval, {MIPR} 2019,
  San Jose, CA, USA, March 28-30, 2019}.\hskip 1em plus 0.5em minus 0.4em\relax
  {IEEE}, 2019, pp. 65--70. [Online]. Available:
  \url{https://doi.org/10.1109/MIPR.2019.00020}
\BIBentrySTDinterwordspacing

\bibitem{poem}
\BIBentryALTinterwordspacing
X.~Yi, R.~Li, and M.~Sun, ``Generating chinese classical poems with {RNN}
  encoder-decoder,'' in \emph{Chinese Computational Linguistics and Natural
  Language Processing Based on Naturally Annotated Big Data - 16th China
  National Conference, {CCL} 2017, - and - 5th International Symposium,
  {NLP-NABD} 2017, Nanjing, China, October 13-15, 2017, Proceedings}, ser.
  Lecture Notes in Computer Science, M.~Sun, X.~Wang, B.~Chang, and D.~Xiong,
  Eds., vol. 10565.\hskip 1em plus 0.5em minus 0.4em\relax Springer, 2017, pp.
  211--223. [Online]. Available:
  \url{https://doi.org/10.1007/978-3-319-69005-6\_18}
\BIBentrySTDinterwordspacing

\bibitem{ELMO}
\BIBentryALTinterwordspacing
M.~E. Peters, M.~Neumann, M.~Iyyer, M.~Gardner, C.~Clark, K.~Lee, and
  L.~Zettlemoyer, ``Deep contextualized word representations,'' in
  \emph{Proceedings of the 2018 Conference of the North American Chapter of the
  Association for Computational Linguistics: Human Language Technologies,
  {NAACL-HLT} 2018, New Orleans, Louisiana, USA, June 1-6, 2018, Volume 1 (Long
  Papers)}, M.~A. Walker, H.~Ji, and A.~Stent, Eds.\hskip 1em plus 0.5em minus
  0.4em\relax Association for Computational Linguistics, 2018, pp. 2227--2237.
  [Online]. Available: \url{https://doi.org/10.18653/v1/n18-1202}
\BIBentrySTDinterwordspacing

\bibitem{GPT}
A.~Radford, K.~Narasimhan, T.~Salimans, and I.~Sutskever, ``Improving language
  understanding by generative pre-training,'' 2018.

\bibitem{BERT}
\BIBentryALTinterwordspacing
J.~Devlin, M.~Chang, K.~Lee, and K.~Toutanova, ``{BERT:} pre-training of deep
  bidirectional transformers for language understanding,'' in \emph{Proceedings
  of the 2019 Conference of the North American Chapter of the Association for
  Computational Linguistics: Human Language Technologies, {NAACL-HLT} 2019,
  Minneapolis, MN, USA, June 2-7, 2019, Volume 1 (Long and Short Papers)},
  J.~Burstein, C.~Doran, and T.~Solorio, Eds.\hskip 1em plus 0.5em minus
  0.4em\relax Association for Computational Linguistics, 2019, pp. 4171--4186.
  [Online]. Available: \url{https://doi.org/10.18653/v1/n19-1423}
\BIBentrySTDinterwordspacing

\bibitem{lei2021have}
W.~Lei, Y.~Miao, R.~Xie, B.~Webber, M.~Liu, T.-S. Chua, and N.~F. Chen, ``Have
  we solved the hard problem? it’s not easy! contextual lexical contrast as a
  means to probe neural coherence,'' in \emph{Proceedings of the AAAI
  Conference on Artificial Intelligence}, 2021.

\bibitem{lei-etal-2020-examining}
\BIBentryALTinterwordspacing
W.~Lei, W.~Wang, Z.~Ma, T.~Gan, W.~Lu, M.-Y. Kan, and T.-S. Chua,
  ``Re-examining the role of schema linking in text-to-{SQL},'' in
  \emph{Proceedings of the 2020 Conference on Empirical Methods in Natural
  Language Processing (EMNLP)}.\hskip 1em plus 0.5em minus 0.4em\relax Online:
  Association for Computational Linguistics, Nov. 2020, pp. 6943--6954.
  [Online]. Available:
  \url{https://www.aclweb.org/anthology/2020.emnlp-main.564}
\BIBentrySTDinterwordspacing

\bibitem{WO2VEC}
\BIBentryALTinterwordspacing
T.~Mikolov, K.~Chen, G.~Corrado, and J.~Dean, ``Efficient estimation of word
  representations in vector space,'' in \emph{1st International Conference on
  Learning Representations, {ICLR} 2013, Scottsdale, Arizona, USA, May 2-4,
  2013, Workshop Track Proceedings}, Y.~Bengio and Y.~LeCun, Eds., 2013.
  [Online]. Available: \url{http://arxiv.org/abs/1301.3781}
\BIBentrySTDinterwordspacing

\bibitem{glove}
\BIBentryALTinterwordspacing
J.~Pennington, R.~Socher, and C.~D. Manning, ``Glove: Global vectors for word
  representation,'' in \emph{Proceedings of the 2014 Conference on Empirical
  Methods in Natural Language Processing, {EMNLP} 2014, October 25-29, 2014,
  Doha, Qatar, {A} meeting of SIGDAT, a Special Interest Group of the {ACL}},
  A.~Moschitti, B.~Pang, and W.~Daelemans, Eds.\hskip 1em plus 0.5em minus
  0.4em\relax {ACL}, 2014, pp. 1532--1543. [Online]. Available:
  \url{https://doi.org/10.3115/v1/d14-1162}
\BIBentrySTDinterwordspacing

\bibitem{XLNet}
\BIBentryALTinterwordspacing
Z.~Yang, Z.~Dai, Y.~Yang, J.~G. Carbonell, R.~Salakhutdinov, and Q.~V. Le,
  ``Xlnet: Generalized autoregressive pretraining for language understanding,''
  in \emph{Advances in Neural Information Processing Systems 32: Annual
  Conference on Neural Information Processing Systems 2019, NeurIPS 2019, 8-14
  December 2019, Vancouver, BC, Canada}, H.~M. Wallach, H.~Larochelle,
  A.~Beygelzimer, F.~d'Alch{\'{e}}{-}Buc, E.~B. Fox, and R.~Garnett, Eds.,
  2019, pp. 5754--5764. [Online]. Available:
  \url{http://papers.nips.cc/paper/8812-xlnet-generalized-autoregressive-pretraining-for-language-understanding}
\BIBentrySTDinterwordspacing

\bibitem{RoBERTa}
\BIBentryALTinterwordspacing
Y.~Liu, M.~Ott, N.~Goyal, J.~Du, M.~Joshi, D.~Chen, O.~Levy, M.~Lewis,
  L.~Zettlemoyer, and V.~Stoyanov, ``Roberta: {A} robustly optimized {BERT}
  pretraining approach,'' \emph{CoRR}, vol. abs/1907.11692, 2019. [Online].
  Available: \url{http://arxiv.org/abs/1907.11692}
\BIBentrySTDinterwordspacing

\bibitem{ALBERT}
\BIBentryALTinterwordspacing
Z.~Lan, M.~Chen, S.~Goodman, K.~Gimpel, P.~Sharma, and R.~Soricut, ``{ALBERT:}
  {A} lite {BERT} for self-supervised learning of language representations,''
  in \emph{8th International Conference on Learning Representations, {ICLR}
  2020, Addis Ababa, Ethiopia, April 26-30, 2020}.\hskip 1em plus 0.5em minus
  0.4em\relax OpenReview.net, 2020. [Online]. Available:
  \url{https://openreview.net/forum?id=H1eA7AEtvS}
\BIBentrySTDinterwordspacing

\bibitem{BioBERT}
\BIBentryALTinterwordspacing
J.~Lee, W.~Yoon, S.~Kim, D.~Kim, S.~Kim, C.~H. So, and J.~Kang, ``Biobert: a
  pre-trained biomedical language representation model for biomedical text
  mining,'' \emph{Bioinform.}, vol.~36, no.~4, pp. 1234--1240, 2020. [Online].
  Available: \url{https://doi.org/10.1093/bioinformatics/btz682}
\BIBentrySTDinterwordspacing

\bibitem{SCIBERT}
\BIBentryALTinterwordspacing
I.~Beltagy, K.~Lo, and A.~Cohan, ``Scibert: {A} pretrained language model for
  scientific text,'' in \emph{Proceedings of the 2019 Conference on Empirical
  Methods in Natural Language Processing and the 9th International Joint
  Conference on Natural Language Processing, {EMNLP-IJCNLP} 2019, Hong Kong,
  China, November 3-7, 2019}, K.~Inui, J.~Jiang, V.~Ng, and X.~Wan, Eds.\hskip
  1em plus 0.5em minus 0.4em\relax Association for Computational Linguistics,
  2019, pp. 3613--3618. [Online]. Available:
  \url{https://doi.org/10.18653/v1/D19-1371}
\BIBentrySTDinterwordspacing

\bibitem{alsentzer-etal-2019-publicly}
\BIBentryALTinterwordspacing
E.~Alsentzer, J.~Murphy, W.~Boag, W.-H. Weng, D.~Jindi, T.~Naumann, and
  M.~McDermott, ``Publicly available clinical {BERT} embeddings,'' in
  \emph{Proceedings of the 2nd Clinical Natural Language Processing
  Workshop}.\hskip 1em plus 0.5em minus 0.4em\relax Minneapolis, Minnesota,
  USA: Association for Computational Linguistics, Jun. 2019, pp. 72--78.
  [Online]. Available: \url{https://www.aclweb.org/anthology/W19-1909}
\BIBentrySTDinterwordspacing

\bibitem{FlauBERT}
\BIBentryALTinterwordspacing
H.~Le, L.~Vial, J.~Frej, V.~Segonne, M.~Coavoux, B.~Lecouteux, A.~Allauzen,
  B.~Crabb{\'{e}}, L.~Besacier, and D.~Schwab, ``Flaubert: Unsupervised
  language model pre-training for french,'' in \emph{Proceedings of The 12th
  Language Resources and Evaluation Conference, {LREC} 2020, Marseille, France,
  May 11-16, 2020}, N.~Calzolari, F.~B{\'{e}}chet, P.~Blache, K.~Choukri,
  C.~Cieri, T.~Declerck, S.~Goggi, H.~Isahara, B.~Maegaard, J.~Mariani,
  H.~Mazo, A.~Moreno, J.~Odijk, and S.~Piperidis, Eds.\hskip 1em plus 0.5em
  minus 0.4em\relax European Language Resources Association, 2020, pp.
  2479--2490. [Online]. Available:
  \url{https://www.aclweb.org/anthology/2020.lrec-1.302/}
\BIBentrySTDinterwordspacing

\bibitem{camembert}
\BIBentryALTinterwordspacing
L.~Martin, B.~M{\"{u}}ller, P.~J.~O. Su{\'{a}}rez, Y.~Dupont, L.~Romary,
  {\'{E}}.~de~la Clergerie, D.~Seddah, and B.~Sagot, ``Camembert: a tasty
  french language model,'' in \emph{Proceedings of the 58th Annual Meeting of
  the Association for Computational Linguistics, {ACL} 2020, Online, July 5-10,
  2020}, D.~Jurafsky, J.~Chai, N.~Schluter, and J.~R. Tetreault, Eds.\hskip 1em
  plus 0.5em minus 0.4em\relax Association for Computational Linguistics, 2020,
  pp. 7203--7219. [Online]. Available:
  \url{https://www.aclweb.org/anthology/2020.acl-main.645/}
\BIBentrySTDinterwordspacing

\bibitem{BERTje}
\BIBentryALTinterwordspacing
W.~de~Vries, A.~van Cranenburgh, A.~Bisazza, T.~Caselli, G.~van Noord, and
  M.~Nissim, ``Bertje: {A} dutch {BERT} model,'' \emph{CoRR}, vol.
  abs/1912.09582, 2019. [Online]. Available:
  \url{http://arxiv.org/abs/1912.09582}
\BIBentrySTDinterwordspacing

\bibitem{RobBERT}
\BIBentryALTinterwordspacing
P.~Delobelle, T.~Winters, and B.~Berendt, ``Robbert: a dutch roberta-based
  language model,'' \emph{CoRR}, vol. abs/2001.06286, 2020. [Online].
  Available: \url{https://arxiv.org/abs/2001.06286}
\BIBentrySTDinterwordspacing

\bibitem{AraBERT}
\BIBentryALTinterwordspacing
W.~Antoun, F.~Baly, and H.~M. Hajj, ``Arabert: Transformer-based model for
  arabic language understanding,'' \emph{CoRR}, vol. abs/2003.00104, 2020.
  [Online]. Available: \url{https://arxiv.org/abs/2003.00104}
\BIBentrySTDinterwordspacing

\bibitem{poem-template2}
\BIBentryALTinterwordspacing
N.~Tosa, H.~Obara, and M.~Minoh, ``Hitch haiku: An interactive supporting
  system for composing haiku poem,'' in \emph{Entertainment Computing - {ICEC}
  2008, 7th International Conference, Pittsburgh, PA, USA, September 25-27,
  2008. Proceedings}, ser. Lecture Notes in Computer Science, S.~M. Stevens and
  S.~J. Saldamarco, Eds., vol. 5309.\hskip 1em plus 0.5em minus 0.4em\relax
  Springer, 2008, pp. 209--216. [Online]. Available:
  \url{https://doi.org/10.1007/978-3-540-89222-9\_26}
\BIBentrySTDinterwordspacing

\bibitem{kexin}
\BIBentryALTinterwordspacing
K.~Yang, D.~Liu, Q.~Qu, Y.~Sang, and J.~Lv, ``An automatic evaluation metric
  for ancient-modern chinese translation,'' \emph{Neural Comput. Appl.},
  vol.~33, no.~8, pp. 3855--3867, 2021. [Online]. Available:
  \url{https://doi.org/10.1007/s00521-020-05216-8}
\BIBentrySTDinterwordspacing

\bibitem{poem-template3}
\BIBentryALTinterwordspacing
X.~Wu, N.~Tosa, and R.~Nakatsu, ``New hitch haiku: An interactive renku poem
  composition supporting tool applied for sightseeing navigation system,'' in
  \emph{Entertainment Computing - {ICEC} 2009, 8th International Conference,
  Paris, France, September 3-5, 2009. Proceedings}, ser. Lecture Notes in
  Computer Science, S.~Natkin and J.~Dupire, Eds., vol. 5709.\hskip 1em plus
  0.5em minus 0.4em\relax Springer, 2009, pp. 191--196. [Online]. Available:
  \url{https://doi.org/10.1007/978-3-642-04052-8\_19}
\BIBentrySTDinterwordspacing

\bibitem{poem-template1}
\BIBentryALTinterwordspacing
R.~Manurung, G.~Ritchie, and H.~Thompson, ``Using genetic algorithms to create
  meaningful poetic text,'' \emph{J. Exp. Theor. Artif. Intell.}, vol.~24,
  no.~1, pp. 43--64, 2012. [Online]. Available:
  \url{https://doi.org/10.1080/0952813X.2010.539029}
\BIBentrySTDinterwordspacing

\bibitem{poem-neural3}
\BIBentryALTinterwordspacing
Q.~Wang, T.~Luo, and D.~Wang, ``Can machine generate traditional chinese
  poetry? {A} feigenbaum test,'' in \emph{Advances in Brain Inspired Cognitive
  Systems - 8th International Conference, {BICS} 2016, Beijing, China, November
  28-30, 2016, Proceedings}, ser. Lecture Notes in Computer Science, C.~Liu,
  A.~Hussain, B.~Luo, K.~C. Tan, Y.~Zeng, and Z.~Zhang, Eds., vol. 10023, 2016,
  pp. 34--46. [Online]. Available:
  \url{https://doi.org/10.1007/978-3-319-49685-6\_4}
\BIBentrySTDinterwordspacing

\bibitem{poem-neural1}
\BIBentryALTinterwordspacing
X.~Yi, R.~Li, and M.~Sun, ``Generating chinese classical poems with {RNN}
  encoder-decoder,'' in \emph{Chinese Computational Linguistics and Natural
  Language Processing Based on Naturally Annotated Big Data - 16th China
  National Conference, {CCL} 2017, - and - 5th International Symposium,
  {NLP-NABD} 2017, Nanjing, China, October 13-15, 2017, Proceedings}, ser.
  Lecture Notes in Computer Science, M.~Sun, X.~Wang, B.~Chang, and D.~Xiong,
  Eds., vol. 10565.\hskip 1em plus 0.5em minus 0.4em\relax Springer, 2017, pp.
  211--223. [Online]. Available:
  \url{https://doi.org/10.1007/978-3-319-69005-6\_18}
\BIBentrySTDinterwordspacing

\bibitem{poem-neural2}
\BIBentryALTinterwordspacing
D.~Liu, Q.~Guo, W.~Li, and J.~Lv, ``A multi-modal chinese poetry generation
  model,'' in \emph{2018 International Joint Conference on Neural Networks,
  {IJCNN} 2018, Rio de Janeiro, Brazil, July 8-13, 2018}.\hskip 1em plus 0.5em
  minus 0.4em\relax {IEEE}, 2018, pp. 1--8. [Online]. Available:
  \url{https://doi.org/10.1109/IJCNN.2018.8489579}
\BIBentrySTDinterwordspacing

\bibitem{lei-etal-2018-sequicity}
\BIBentryALTinterwordspacing
W.~Lei, X.~Jin, M.-Y. Kan, Z.~Ren, X.~He, and D.~Yin, ``{S}equicity:
  Simplifying task-oriented dialogue systems with single sequence-to-sequence
  architectures,'' in \emph{Proceedings of the 56th Annual Meeting of the
  Association for Computational Linguistics (Volume 1: Long Papers)}.\hskip 1em
  plus 0.5em minus 0.4em\relax Melbourne, Australia: Association for
  Computational Linguistics, Jul. 2018, pp. 1437--1447. [Online]. Available:
  \url{https://www.aclweb.org/anthology/P18-1133}
\BIBentrySTDinterwordspacing

\bibitem{lei22}
\BIBentryALTinterwordspacing
H.~Liang, W.~Lei, P.~Y. Chan, Z.~Yang, M.~Sun, and T.-S. Chua, ``Pirhdy:
  Learning pitch-, rhythm-, and dynamics-aware embeddings for symbolic
  music.''\hskip 1em plus 0.5em minus 0.4em\relax New York, NY, USA:
  Association for Computing Machinery, 2020. [Online]. Available:
  \url{https://doi.org/10.1145/3394171.3414032}
\BIBentrySTDinterwordspacing

\bibitem{poem-rl}
\BIBentryALTinterwordspacing
X.~Yi, M.~Sun, R.~Li, and W.~Li, ``Automatic poetry generation with mutual
  reinforcement learning,'' in \emph{Proceedings of the 2018 Conference on
  Empirical Methods in Natural Language Processing, Brussels, Belgium, October
  31 - November 4, 2018}, E.~Riloff, D.~Chiang, J.~Hockenmaier, and J.~Tsujii,
  Eds.\hskip 1em plus 0.5em minus 0.4em\relax Association for Computational
  Linguistics, 2018, pp. 3143--3153. [Online]. Available:
  \url{https://doi.org/10.18653/v1/d18-1353}
\BIBentrySTDinterwordspacing

\bibitem{poem-VAE}
\BIBentryALTinterwordspacing
X.~Yang, X.~Lin, S.~Suo, and M.~Li, ``Generating thematic chinese poetry using
  conditional variational autoencoders with hybrid decoders,'' in
  \emph{Proceedings of the Twenty-Seventh International Joint Conference on
  Artificial Intelligence, {IJCAI} 2018, July 13-19, 2018, Stockholm, Sweden},
  J.~Lang, Ed.\hskip 1em plus 0.5em minus 0.4em\relax ijcai.org, 2018, pp.
  4539--4545. [Online]. Available:
  \url{https://doi.org/10.24963/ijcai.2018/631}
\BIBentrySTDinterwordspacing

\bibitem{couplet-smt}
\BIBentryALTinterwordspacing
L.~Jiang and M.~Zhou, ``Generating chinese couplets using a statistical {MT}
  approach,'' in \emph{{COLING} 2008, 22nd International Conference on
  Computational Linguistics, Proceedings of the Conference, 18-22 August 2008,
  Manchester, {UK}}, D.~Scott and H.~Uszkoreit, Eds., 2008, pp. 377--384.
  [Online]. Available: \url{https://www.aclweb.org/anthology/C08-1048/}
\BIBentrySTDinterwordspacing

\bibitem{Transformer}
\BIBentryALTinterwordspacing
A.~Vaswani, N.~Shazeer, N.~Parmar, J.~Uszkoreit, L.~Jones, A.~N. Gomez,
  L.~Kaiser, and I.~Polosukhin, ``Attention is all you need,'' in
  \emph{Advances in Neural Information Processing Systems 30: Annual Conference
  on Neural Information Processing Systems 2017, 4-9 December 2017, Long Beach,
  CA, {USA}}, I.~Guyon, U.~von Luxburg, S.~Bengio, H.~M. Wallach, R.~Fergus,
  S.~V.~N. Vishwanathan, and R.~Garnett, Eds., 2017, pp. 5998--6008. [Online].
  Available: \url{http://papers.nips.cc/paper/7181-attention-is-all-you-need}
\BIBentrySTDinterwordspacing

\bibitem{adam}
\BIBentryALTinterwordspacing
D.~P. Kingma and J.~Ba, ``Adam: {A} method for stochastic optimization,'' in
  \emph{3rd International Conference on Learning Representations, {ICLR} 2015,
  San Diego, CA, USA, May 7-9, 2015, Conference Track Proceedings}, Y.~Bengio
  and Y.~LeCun, Eds., 2015. [Online]. Available:
  \url{http://arxiv.org/abs/1412.6980}
\BIBentrySTDinterwordspacing

\bibitem{Wolf2019HuggingFacesTS}
T.~Wolf, L.~Debut, V.~Sanh, J.~Chaumond, C.~Delangue, A.~Moi, P.~Cistac,
  T.~Rault, R.~Louf, M.~Funtowicz, and J.~Brew, ``Huggingface's transformers:
  State-of-the-art natural language processing,'' \emph{ArXiv}, vol.
  abs/1910.03771, 2019.

\bibitem{BLEU}
\BIBentryALTinterwordspacing
K.~Papineni, S.~Roukos, T.~Ward, and W.~Zhu, ``Bleu: a method for automatic
  evaluation of machine translation,'' in \emph{Proceedings of the 40th Annual
  Meeting of the Association for Computational Linguistics, July 6-12, 2002,
  Philadelphia, PA, {USA}}.\hskip 1em plus 0.5em minus 0.4em\relax {ACL}, 2002,
  pp. 311--318. [Online]. Available:
  \url{https://www.aclweb.org/anthology/P02-1040/}
\BIBentrySTDinterwordspacing

\bibitem{lei2018sequicity}
W.~Lei, X.~Jin, M.-Y. Kan, Z.~Ren, X.~He, and D.~Yin, ``Sequicity: Simplifying
  task-oriented dialogue systems with single sequence-to-sequence
  architectures,'' in \emph{Proceedings of the 56th Annual Meeting of the
  Association for Computational Linguistics (Volume 1: Long Papers)}, 2018, pp.
  1437--1447.

\bibitem{jin2018explicit}
X.~Jin, W.~Lei, Z.~Ren, H.~Chen, S.~Liang, Y.~Zhao, and D.~Yin, ``Explicit
  state tracking with semi-supervisionfor neural dialogue generation,'' in
  \emph{Proceedings of the 27th ACM International Conference on Information and
  Knowledge Management}, 2018, pp. 1403--1412.

\bibitem{liao2020rethinking}
L.~Liao, Y.~Ma, W.~Lei, and T.-S. Chua, ``Rethinking dialogue state tracking
  with reasoning,'' \emph{arXiv preprint arXiv:2005.13129}, 2020.

\bibitem{SeqGAN}
\BIBentryALTinterwordspacing
L.~Yu, W.~Zhang, J.~Wang, and Y.~Yu, ``Seqgan: Sequence generative adversarial
  nets with policy gradient,'' in \emph{Proceedings of the Thirty-First {AAAI}
  Conference on Artificial Intelligence, February 4-9, 2017, San Francisco,
  California, {USA}}, S.~P. Singh and S.~Markovitch, Eds.\hskip 1em plus 0.5em
  minus 0.4em\relax {AAAI} Press, 2017, pp. 2852--2858. [Online]. Available:
  \url{http://aaai.org/ocs/index.php/AAAI/AAAI17/paper/view/14344}
\BIBentrySTDinterwordspacing

\end{thebibliography}

\end{document}